\documentclass{article}

\usepackage{microtype}
\usepackage{graphicx}
\usepackage{subfigure}
\usepackage{booktabs} 

\usepackage{hyperref}

\usepackage[accepted]{icml2024}

\usepackage{amsmath}
\usepackage{amssymb}
\usepackage{mathtools}
\usepackage{amsthm}
\usepackage{MnSymbol}

\usepackage[capitalize,noabbrev]{cleveref}

\theoremstyle{plain}

\theoremstyle{definition}

\theoremstyle{remark}

\usepackage[textsize=tiny]{todonotes}

\usepackage{array}          
\usepackage{booktabs}       
\usepackage{multirow} 
\usepackage{amsfonts}       
\usepackage{xcolor}         
\usepackage{wrapfig}        
\usepackage{makecell}       
\usepackage[most]{tcolorbox}

\usepackage{listings}
\usepackage{xcolor}

\definecolor{codegreen}{rgb}{0,0.6,0}
\definecolor{codegray}{rgb}{0.5,0.5,0.5}
\definecolor{codepurple}{rgb}{0.58,0,0.82}
\definecolor{backcolour}{rgb}{0.95,0.95,0.92}

\lstdefinestyle{mystyle}{
    backgroundcolor=\color{backcolour},   
    commentstyle=\color{codegreen},
    keywordstyle=\color{magenta},
    numberstyle=\tiny\color{codegray},
    stringstyle=\color{codepurple},
    basicstyle=\ttfamily\footnotesize,
    breakatwhitespace=false,         
    breaklines=true,                 
    captionpos=b,                    
    keepspaces=true,                 
    numbers=left,                    
    numbersep=5pt,                  
    showspaces=false,                
    showstringspaces=false,
    showtabs=false,                  
    tabsize=2
}
\usepackage{amsmath,amsfonts,bm}

\newtcolorbox[auto counter]{Summary}[1][]{title={\bfseries Pitfall~\thetcbcounter},enhanced,drop shadow={black!50!white},
  coltitle=black,
  top=0.13in,
  attach boxed title to top left=
  {xshift=1.5em,yshift=-\tcboxedtitleheight/2},
  boxed title style={size=small,colback=white},#1}

\newcommand{\fref}[1]{Figure~\ref{#1}}

\newcommand{\tref}[1]{Table~\ref{#1}}

\newcommand{\sref}[1]{$\S$~\ref{#1}}
\newcommand{\aref}[1]{Appendix \ref{#1}}

\newcommand\modify[1]{\textcolor{black}{#1}}
\newcommand\yh[1]{\textcolor{black}{#1}}
\newcommand\jo[1]{\textcolor{black}{#1}}
\newcommand\jjo[1]{\textcolor{black}{#1}}

\newcommand\jh[1]{\textcolor{black}{#1}}

\def\v_GT{v_{\text{GT}}}

\def\tx{\tilde{x}}

\def\predictedx0{x_0}

\icmltitlerunning{GOAR: Coordinate-Invariant eXplainable AI Assessment}

\begin{document}

\twocolumn[
\icmltitle{
    Geometric Remove-and-Retrain (GOAR):\\ 
    Coordinate-Invariant eXplainable AI Assessment
}



\icmlsetsymbol{equal}{*}

\begin{icmlauthorlist}
\icmlauthor{Yong-Hyun Park}{snu}
\icmlauthor{Junghoon Seo}{si}
\icmlauthor{Bomseok Park}{snu}
\icmlauthor{Seongsu Lee}{snu}
\icmlauthor{Junghyo Jo}{snu}
\end{icmlauthorlist}

\icmlaffiliation{snu}{Department of Physics Education, Seoul Nataional University}
\icmlaffiliation{si}{SI-Analytics}

\icmlcorrespondingauthor{Yong-Hyun Park}{enkeejunior1@snu.ac.kr}

\icmlkeywords{Machine Learning, ICML}

\vskip 0.3in
]



\printAffiliationsAndNotice{}  

\begin{abstract}
Identifying the relevant input features that have a critical influence on the output results is indispensable for the development of explainable artificial intelligence (XAI).
Remove-and-Retrain (ROAR) is a widely accepted approach for assessing the importance of individual pixels by measuring changes in accuracy following their removal and subsequent retraining of the modified dataset. 
However, we uncover notable limitations in pixel-perturbation strategies. 
\yh{When viewed from a geometric perspective, we discover that these metrics fail to discriminate between differences among feature attribution methods, thereby compromising the reliability of the evaluation.}
To address this challenge, we introduce an alternative feature-perturbation approach named Geometric Remove-and-Retrain (GOAR). 
Through a series of experiments with both synthetic and real datasets, we substantiate that GOAR transcends the limitations of pixel-centric metrics. 
\end{abstract}

\section{Introduction}

\begin{figure}[!t]
    \centering
    \includegraphics[width=1\linewidth]{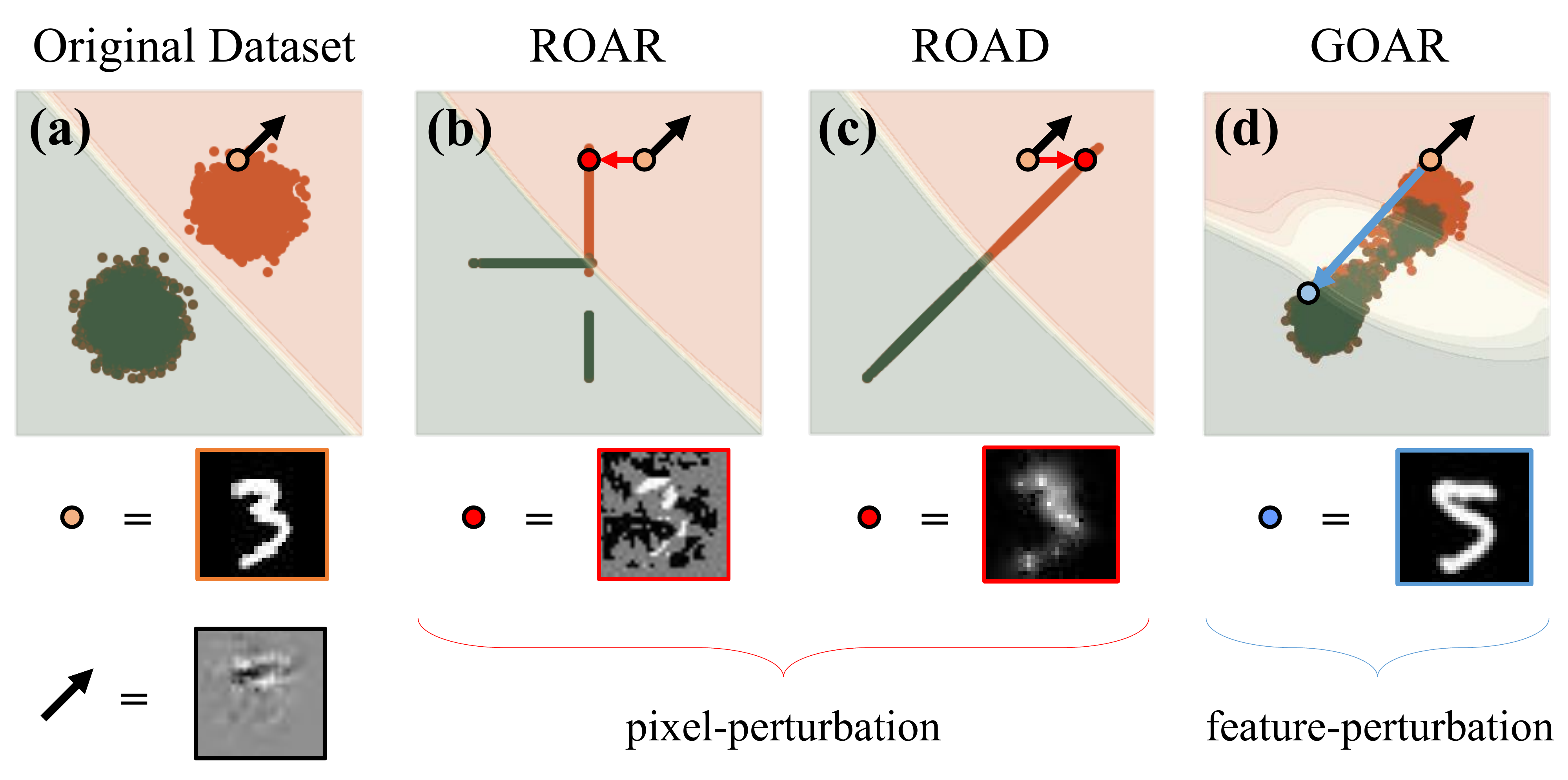}
    \vspace{-2mm}
    \caption{
    \textbf{\yh{Comparison between previous perturbation strategies (ROAR, ROAD) and ours (GOAR).}}
    The perturbation-based metric evaluates the significance of features by comparing the model's performance between the original dataset (a) and a modified dataset where significant features have been removed (b, c, d). The top row illustrates the feature removal process with a 2D dataset, while the bottom row visualizes the samples after removing features based on each benchmark's perturbation method. The shaded regions in the 2D dataset represent the decision boundaries of models trained on each dataset.
    (a) The black arrow indicates the feature discovered using the attribution method for the sample marked with a circle.
    (b, c) Pixel-perturbation methods like Remove-and-Retrain (ROAR) and Remove-and-Debias (ROAD) fail to eliminate important information from samples, resulting in little degradation in performance.
    This is because these methods move samples in a pixel-basis direction (red arrow) unrelated to the feature direction (black arrow) when erasing information.
    (d) In contrast, our GOAR offers a precise and effective way to remove features, enabling an accurate comparison of attribution methods.
    }
    \vspace{-2mm}
    \label{fig:teaser}
\end{figure}

Feature attribution methods, \jo{which involve the identification of input features} most relevant to a model's output, are pivotal in explainable artificial intelligence (XAI) research \citep{simonyan2014deep, shrikumar2017just, smilkov2017smoothgrad, sundararajan2017axiomatic, rong2022consistent}.
These explanations have been used to both analyze \citep{magesh2020explainable,mirzavand2023explainable,zeineldin2022explainability} and debug \citep{bhatt2020evaluating, adebayo2020debugging} neural networks.
As various attribution methods are being proposed, it becomes essential to set up benchmarks to evaluate them \citep{nguyen2020quantitative, hase2020evaluating, yeh2019infidelity, hooker2019benchmark}.

\jo{A widely used approach for assessing attribution methods is the \textit{pixel-perturbation} strategy}.
This strategy removes pixels associated with relevant features from an image and then measures the model performance after retraining on this modified dataset. 
If the model's accuracy drops significantly, it indicates the method successfully identified crucial features.
Within this strategy, \jo{two prominent metrics stand out}: ``Remove-and-Retrain'' (ROAR)~\citep{hooker2019benchmark}, which replaces important pixels with a fixed value, and ``Remove-and-Debias'' (ROAD)~\citep{rong2022consistent}, which \jo{substitutes} pixels with a noisy linear interpolation.



Despite the widespread use of pixel-perturbation strategies, we reveal a significant \jo{geometric issue associated with} them. 
{
Geometrically, one can think of the perturbation as moving each sample in space. 
However, the pixel-perturbation strategy, i.e., moving in the pixel-basis direction, cannot effectively erase the information contained in the dataset.
}

{
To elucidate this issue, \jo{we} consider a 2-class Gaussian mixture dataset (\fref{fig:teaser}). 
In this example, 
\yh{pixel-based perturbations (ROAR and ROAD)} fail to effectively erase information, leaving the dataset still separable. 
Neither good nor bad features cannot lead to an accuracy drop after pixel-perturbation, resulting in an ineffective comparison of features.
}

To address this problem, we introduce \jo{a \textit{feature-perturbation} strategy named} Geometric Remove-and-Retrain (GOAR). 
GOAR perturbs images by moving the data sample along the feature direction, eliminating the dependence on pixel coordinates. 
However, na\"ively shifting the image along the feature direction can lead to a retrained model predicting class labels based on {spurious patterns related to the subtracted features.}
We tackle this issue by leveraging the diffusion model to ensure that perturbed data points remain closely aligned with their original data manifold.
As shown in \fref{fig:teaser}-(c), unlike other metrics, GOAR removes information from the dataset in a precise and effective way.

We validate our geometric analysis and the effectiveness of GOAR through experiments on both synthetic and real datasets.
First, using synthetic datasets, we show pixel-perturbation strategies fail to discriminate between features, whereas GOAR provides accurate comparisons.
Second, we demonstrate that GOAR aligns well with ground truth feature evaluation results. 
Here, we follow the OpenXAI \cite{agarwal2022openxai} benchmark that provides ground truth features based on interpretable models.
\yh{
Finally, we carry out experiments across a range of datasets to demonstrate the versatility of GOAR in various domains. These datasets include image datasets like MNIST \citep{deng2012mnist}, FashionMNIST \citep{xiao2017fashionmnist}, and CIFAR10 \citep{Krizhevsky2009LearningML}, as well as tabular datasets such as Iris \citep{misc_iris_53}, Raisin \citep{misc_raisin_850}, and Wine \citep{misc_wine_109}. 
This diverse set of datasets highlights how GOAR can effectively perform in different domains.
}




Our contributions are summarized as follows:
\begin{itemize}
  \item We examine pixel-perturbation strategies from a geometric perspective and reveal their limitations arising from their reliance on pixel coordinates.
  \item We propose a new feature-perturbation strategy \jo{named} GOAR.
  \item \yh{We validate the efficacy of GOAR by comparing the evaluation results based on ground truth features.}
  \item We demonstrate the broad applicability of GOAR through experiments on various datasets, spanning across different domains such as vision and tabular data.
\end{itemize}

\section{Related Works}

\paragraph{Feature Attribution Method\jo{s}.}
Feature attribution methods aim to discover the relevant feature of given input that influences the model's prediction behind the scenes. These can be categorized into \jo{(i)} perturbation-based approaches and \jo{(ii)} gradient-based approaches. Perturbation-based approaches \citep{ribeiro2016should, lundberg2017unified} involve removing parts of the input and extracting the features the model focuses on based on their impact. 
Gradient-based approaches \citep{selvaraju2017grad, springenberg2014striving, simonyan2014deep, smilkov2017smoothgrad, sundararajan2017axiomatic, shrikumar2017just, shrikumar2017learning} find features that make the model highly responsive by using concerning to the input or intermediate latent space.
In this study, we use input gradient \citep{simonyan2014deep}, SmoothGrad \citep{smilkov2017smoothgrad}, Integrated gradient \citep{sundararajan2017axiomatic}, Grad $\times$ Input \citep{shrikumar2017just}, DeepLift \citep{shrikumar2017learning}, and SHAP \citep{lundberg2017unified} for the comparison. 

\paragraph{Evaluation Metrics for the Feature Attribution Methods.}

Various evaluation strategies have been proposed so far, such as sanity checks \citep{tomsett2020sanity, adebayo2020debugging}, human explanations \citep{jeyakumar2020can}, perturbation based metric \citep{hooker2019benchmark, jethani2021have, rong2022consistent} or 
\yh{comparing ground truth features \citep{xai-bench-2021, agarwal2022openxai}.} 

\jo{ROAR \citep{hooker2019benchmark} is one of the standard metrics that} evaluates the feature maps based on perturbation. 
\yh{
It involves removing important pixels based on these feature maps and observing the accuracy drop in the model retrained on the modified dataset.
The retraining step is essential to address the out-of-distribution (OOD) problem introduced by pixel masking.
Several works have pointed out issues with ROAR from an information theory perspective and proposed methods to resolve them \citep{jethani2021have, rong2022consistent}. 
In contrast, our work is the first to identify and tackle the geometric challenges associated with the pixel-perturbation process.
}

\yh{
Evaluating attribution methods through ground truth features is also widely used as a benchmark.
Some methods within this category rely on synthetic datasets (XAI-bench) \citep{xai-bench-2021} or interpretable models (OpenXAI) \citep{agarwal2022openxai}.
While this approach excels at precisely assessing features, it has the drawback of being applicable only in very limited scenarios. In our work, we demonstrate that GOAR can yield results similar to ground truth benchmarks while being applicable to a wider range of data and models.
}

\paragraph{{Manifold Projection {via Diffusion Models.}}}
Several methods have been proposed to re-project images that have deviated from the image manifold. 
One notable approach is found in the realm of inverse problems, where the goal is to restore the clean data given degraded observation. 
\yh{Recently, there have been significant advancements in solving inverse problems through diffusion models. These methods are categorized into two main groups: one that relies solely on denoising steps \citep{choi2021ilvr, meng2021sdedit, lugmayr2022repaint, kawar2022denoising} and the other incorporates additional guidance, requiring gradient computations \citep{chung2022improving, chung2022diffusion, chung2023fast}.
In this work, we adopt a first approach to realign perturbed samples on the data manifold.}

\section{Preliminary}

\label{section:background}


\paragraph{Remove-and-Retrain (ROAR) \citep{hooker2019benchmark}}
ROAR operates by removing pixels associated with important features. The performance is then evaluated based on the drop in accuracy when the model is retrained on this modified dataset.
A steeper decline in performance suggests that the attribution method has more accurately identified the critical features \jo{upon which} the model relies for predictions. This concept is referred to as {\it fidelity} \citep{elton2020self}.


To be self-contained, we define the ROAR process as follows: Given a dataset $\mathcal{D} = \{(x_i, y_i)\}_{i=1}^n$, where {$x_i \in \mathbb{R}^n$, and $y_i \in \{1, ..., c\}$}, along with a classifier $f: \mathbb{R}^n \rightarrow \{1, \cdots, c\}$, 
our goal is to assess the attribution methods $e: x_i \mapsto v_i$, where $v_i$ denotes the feature obtained from each sample \jo{$x_i$}. 
Now, \jo{we} choose the pixel with the highest k\% value from $v_i$ and set the corresponding pixels in $x_i$ to a fixed value of 0.
This process can be represented as $\tx_i = (1-M_i) \odot x_i = \pi_i(x_i)$, where $M_i$ is the binary mask where erasing pixels are set to 1, and $\odot$ denotes element-wise multiplication.
Then, we train the model on the modified dataset $\tilde{\mathcal{D}} = \{(\tx_i, y_i)\}_{i=1}^n$ and measure how much the accuracy decreases.

\paragraph{Remove-and-Debias (ROAD) \citep{rong2022consistent}}

After the introduction of ROAR, several studies identified potential pitfalls related to fixed value imputation
\citep{rong2022consistent, song2023pitfalls, jethani2021have}.
When the model is retrained on the modified dataset, it learns to rely on the mask-related information in $\tx_i$ to predict the class label.
This makes the model exploit the spurious correlation between the mask and the label, preventing us from measuring how much information has been lost from $x_i$ due to pixel perturbation.

\citet{rong2022consistent} employ information theory and demonstrate that the issue stems from high mutual information between the modified image $\tx_i$ and the imputation mask $M_i$. 
They define this problem as \textit{class information leakage}.
Based on this analysis, ROAD adopts a different approach for pixel imputation, utilizing noisy linear interpolation instead of fixed values. 
With noisy linear imputation, \jo{ROAD} effectively removes the information of $M_i$ from $\tx_i$, thereby preventing the model from predicting the class label based on mask-related information.
\yh{Since noisy linear imputation is heavily based on the property of image datasets, ROAD is only applicable to the vision domain.}

\paragraph{Eval-X \citep{jethani2021have}} 
\yh{
\citet{jethani2021have} also raises a similar concern and suggests using a proxy model that can withstand the masking operation to measure the accuracy drop. 
To train this proxy model, they incorporate randomly mask pixels with a 50\% probability during the training. 
They named this variation of ROAR, ``Eval-X."
}

\section{\jo{Geometric challenges in ROAR}}

\label{section:theory}

In \sref{subsection:theory_analysis}, we analyze ROAR's evaluation mechanism from a geometric perspective.
In \sref{subsection:theory_pitfalls}, we formalize the pitfalls of ROAR's pixel-perturbation strategy.
In \sref{subsection:theory_origin}, 
we attribute the cause of issues with ROAR to two main factors: the reliance on pixel-coordinate and the all-or-none removal processes.


\subsection{\jo{Mechanism of ROAR}}
\label{subsection:theory_analysis}

We formalize the mechanism of \jo{the} accuracy drop in ROAR. 
For ease of discussion, \jo{we} consider a dataset $\mathcal{D} = \{(x_1, 0), (x_2, 1)\}$ with only one data point for each of the binary labels, \jo{0 or 1}. 
\jo{Note that the input $x_i = (x_i^1, x_i^2, \cdots, x_i^n)$ is an $n$-dimensional vector, where the subscript denotes the sample index, and the superscript indicates vector elements, i.e., pixels.}
We assume the model mispredicts when the difference between \jo{perturbed} $\tx_1$ and $\tx_2$ for all pixels is less than $\epsilon$. 
This corresponds to a situation where $\tx_1$ and $\tx_2$ are too similar to distinguish. 

Upon perturbing $x_1$ and $x_2$ with ROAR (replacing important pixels with zeros), let the difference between them be denoted as $dx = (dx^1, \ldots, dx^n)$.
Among these $dx^i$ values, we define coordinates with a magnitude greater than $\epsilon$ as \textit{relevant}, while the rest as \textit{irrelevant}.
According to our assumption, irrelevant coordinates do not contribute to distinguishing between the two samples, as their differences are already smaller than $\epsilon$.
On the other hand, in the case of relevant coordinates, the information of only a single relevant coordinate is sufficient to make samples separable.
Therefore, the accuracy drop in ROAR occurs when all relevant coordinates are erased for each sample.


\begin{figure*}[!t]
    \centering
\includegraphics[width=1.0\linewidth]{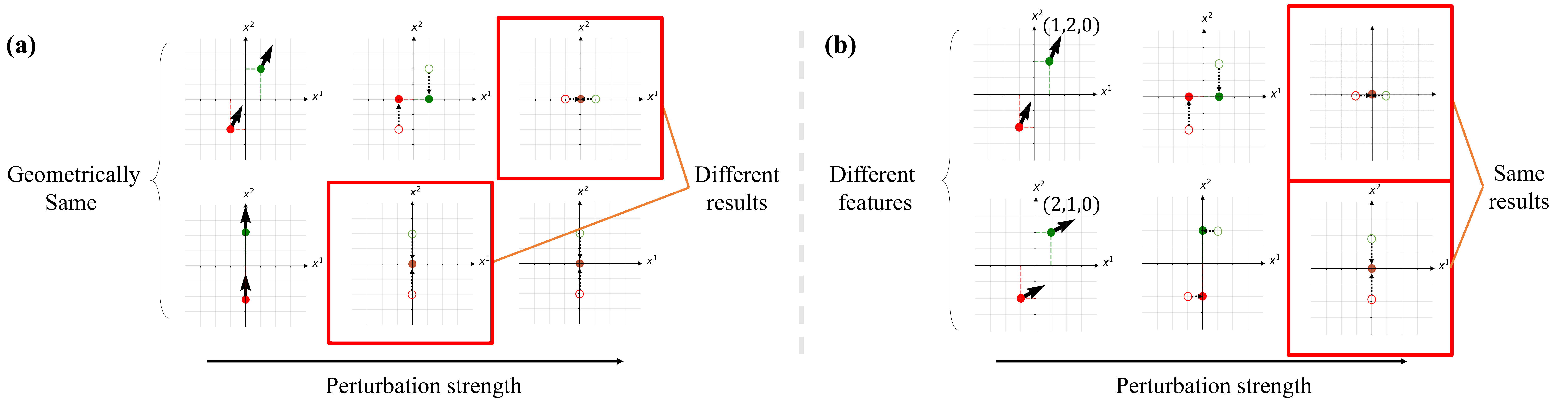}
    \caption{\textbf{Pitfalls of ROAR.}
    Red and Green circles are two data points corresponding to different classes. Black arrows ($\rightarrow$) represent the feature vectors obtained from the attribution method and dashed arrows ($\dashedrightarrow$) 
    indicate the pixel-perturbation based on ROAR. 
    Red borders indicate the points where the performance drop occurs.
    (a) ROAR produces different results in the same geometrical situation depending on coordinate choices.
    (b) ROAR can produce the same results with different features, which makes it challenging to discriminate between these various features.
    }
    \vspace{-1em}
    \label{fig:pitfalls}
\end{figure*}

\subsection{Pitfalls of ROAR}

\label{subsection:theory_pitfalls}

From the findings of \sref{subsection:theory_analysis}, we uncover two significant problems of ROAR. 
\fref{fig:pitfalls} illustrates these issues with minimal examples. 

The first problem is that ROAR yields varying results depending on the choice of coordinates. 
{
To make the discussion easier, let's assume that $v_1 = v_2 = dx$.
In this scenario, ROAR removes pixels from each image in descending order of their corresponding $dx$ values.
Consequently, the drop in accuracy occurs when the number of removed pixels is equal to the number of relevant coordinates.
}

Now, let's consider a different coordinate system where we represent the same dataset with $dx = (dx^1, 0, ..., 0)$. In this new coordinate system, the feature representation becomes $v_1 = v_2 = (1, 0, \cdots, 0)$. As a result, removing just one pixel in this coordinate system leads to performance degradation. 
This example highlights that even when dealing with identical geometric situations, ROAR can lead to different results based on coordinate choices.


\begin{Summary}
    ROAR is {\it not} invariant to the coordinate transformation. 
\end{Summary}

The second problem is that ROAR cannot distinguish differences in the feature vector at relevant coordinates.
To illustrate this, \jo{we} consider a dataset $\mathcal D$ defined by $dx = (1, 2, \epsilon)$, 
\jo{and} two attribution methods, denoted as $e$ and $e'$.
Let's assume features derived from $e$ are $v_1 = v_2 = (1, 2, 0)$, while those from $e'$ are $v'_1 = v'_2 = (2, 1, 0)$.
Ideally, one might anticipate ROAR to favor method $e$ due to its feature alignment with the distinctive direction, i.e., $dx$, of the samples.
However, the accuracy drop in ROAR only occurs after removing all the relevant coordinates, which in this case are the first two pixels. 
Therefore, both $e$ and $e'$ witness an accuracy drop after two pixels are removed, rendering them indistinguishable.
This example highlights that variations in relevant coordinates do not influence ROAR's outcomes, making it unable to distinguish between methods $e$ and $e'$.
Importantly, this limitation indicates that the main purpose of ROAR, feature comparison, might remain unfulfilled.

\begin{Summary}
    ROAR is {\it not} discriminative to differences in features at relevant coordinates.
\end{Summary}

\modify{While pitfall 1 addresses the theoretical issues of ROAR, pitfall 2 pertains to situations that can frequently occur when using pixel-perturbation in practice. We are empirically demonstrating this in \sref{subsec:synthetic}.}

\modify{In the \aref{appendix:pitfalls}, we demonstrate that other pixel-perturbation strategies such as Eval-X and ROAD also suffer the same problems of ROAR.}


\subsection{Origin of Pitfalls}

\label{subsection:theory_origin}

Pitfall 1 arises because ROAR removes features from images based on pixel coordinates, ignoring the data's geometric structure.
Therefore, to mitigate this issue, the pixel-perturbation strategy should be replaced with a \textit{coordinate-free} perturbation approach.


The reason for Pitfall 2 is that when ROAR erases pixel information, it does so in an all-or-none manner. 
{
Because the information in a pixel remains intact until it is removed, it is possible to prevent an accuracy drop by retaining only those pixels that are perfectly intact.
To resolve this issue, we need to \textit{gradually} remove all features from the image.
}

\begin{figure}[!t]
    \centering
    \includegraphics[width=1.0\linewidth]{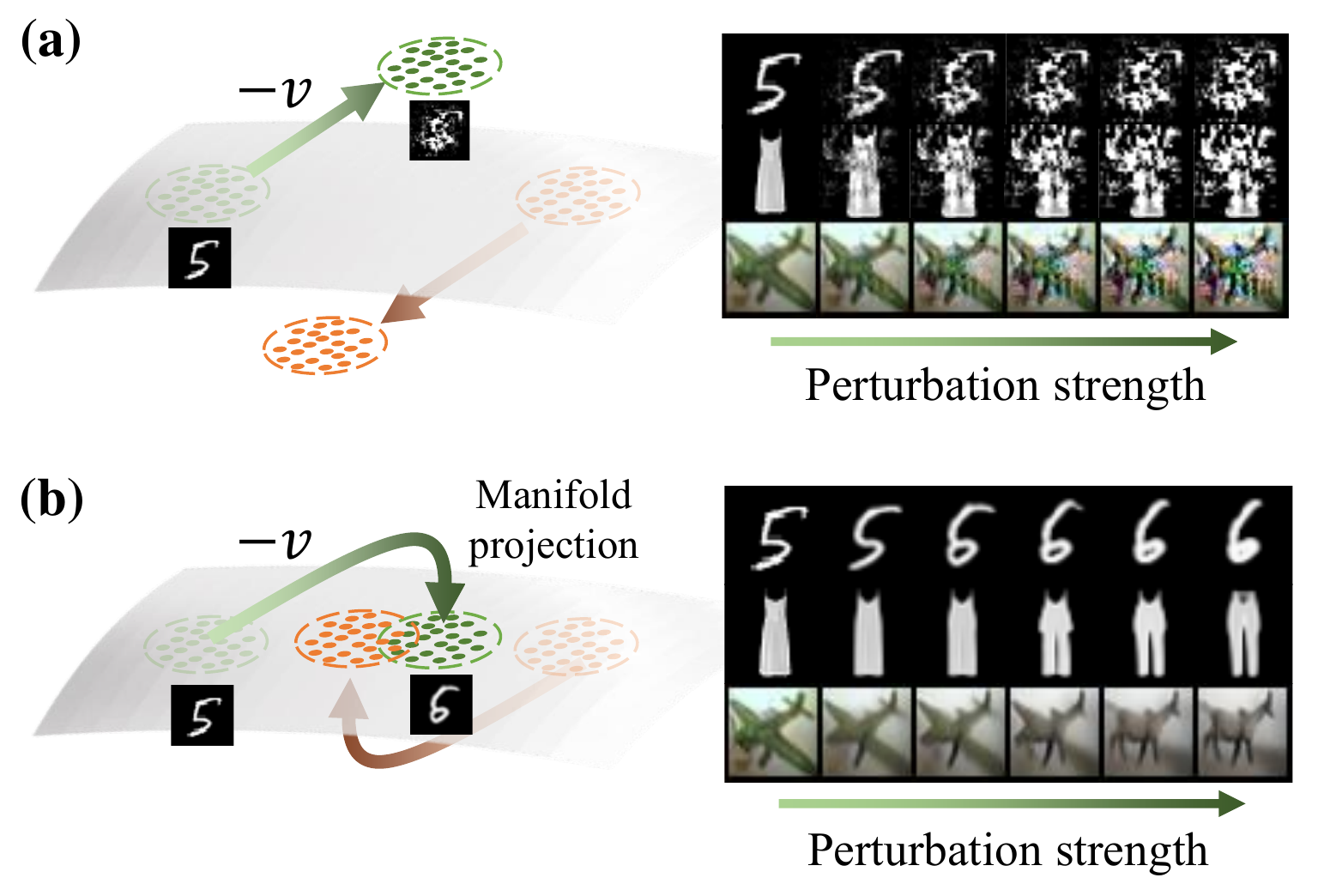}
    \caption{\jo{
    \textbf{Projection onto the data manifold}.
    Perturbation of input data (a) without manifold projection and (b) with manifold projection. 
    }}
    \label{fig:method}
\end{figure}

\section{Geometric Remove-And-Retrain \jo{(GOAR)}}

\label{section:method}

In this section, we introduce a novel evaluation metric using the feature-perturbation strategy: \jo{GOAR}. 
{
In \sref{subsection:before_diffusion_model}, we introduce the feature-perturbation strategy and explain that due to the presence of off-manifold components in the feature vector, na\"ive perturbation along the feature vector is not advisable.
In \sref{subsection:after_diffusion_model}, we present how to remove off-manifold components after perturbation by utilizing the diffusion model.
In \sref{subsection:measure_performance}, we explain how to measure performance degradation in GOAR.
}

\subsection{\jo{Feature Perturbation}}
\label{subsection:before_diffusion_model}

In \sref{section:theory}, our analysis reveals that the pitfalls of ROAR stem from its reliance on pixel coordinates and all-or-none manner removal.
To overcome this drawback, we require a novel evaluation metric that perturbs images without depending on pixel coordinates.

{A straightforward approach is shifting samples in the feature direction, denoted as $\tx_i = x_i - v_i$.} 
However, as depicted in \fref{fig:method}, if the feature vector has an off-manifold component, this approach leads to samples escaping the data manifold \citep{bordt2023manifold,srinivas2023models,srinivas2021rethinking}. 
When we retrain the model on this modified dataset, it tends to predict the class label based on the off-manifold feature in $v_i$ rather than utilizing the remaining information of $x_i$.
This problem can be viewed as a type of class information leakage that arises from significant mutual information between $\tx_i$ and $v_i$. 
To mitigate this problem, it is imperative to eliminate the off-manifold components.


\subsection{\jo{Projection onto the Data Manifold}}

\label{subsection:after_diffusion_model}

To eliminate the off-manifold component from $\tx_i = x_i - v_i$, we need to project the perturbed image onto the data manifold. 
Our goal is to remove any off-manifold components in $\tx_i$ while preserving its on-manifold components. 
\yh{To achieve this, we employed a manifold projection method known as SDEdit \citep{meng2021sdedit}.
SDEdit involves adding a slight amount of Gaussian noise to images outside of the manifold and then denoising them using a diffusion model.
We chose SDEdit because it is the most cost-effective while still delivering satisfactory results.
\jh{As mentioned in Section \ref{sec:discussions},} exploring other manifold projection methods is a subject for future research.}
\jjo{We made slight adaptations to SDEdit to better suit our context. First, instead of employing an SDE sampler, we utilized an ODE sampler \citep{song2021denoising} to avoid alterations in crucial signals from the original images. Additionally, since the feature vector $v$ is often noisy, we treat it as noise when calculating the diffusion timestep $t$. This allows us to effectively remove excessive noise.}

\begin{figure}[!t]
    \centering
    \includegraphics[width=1.0\linewidth]{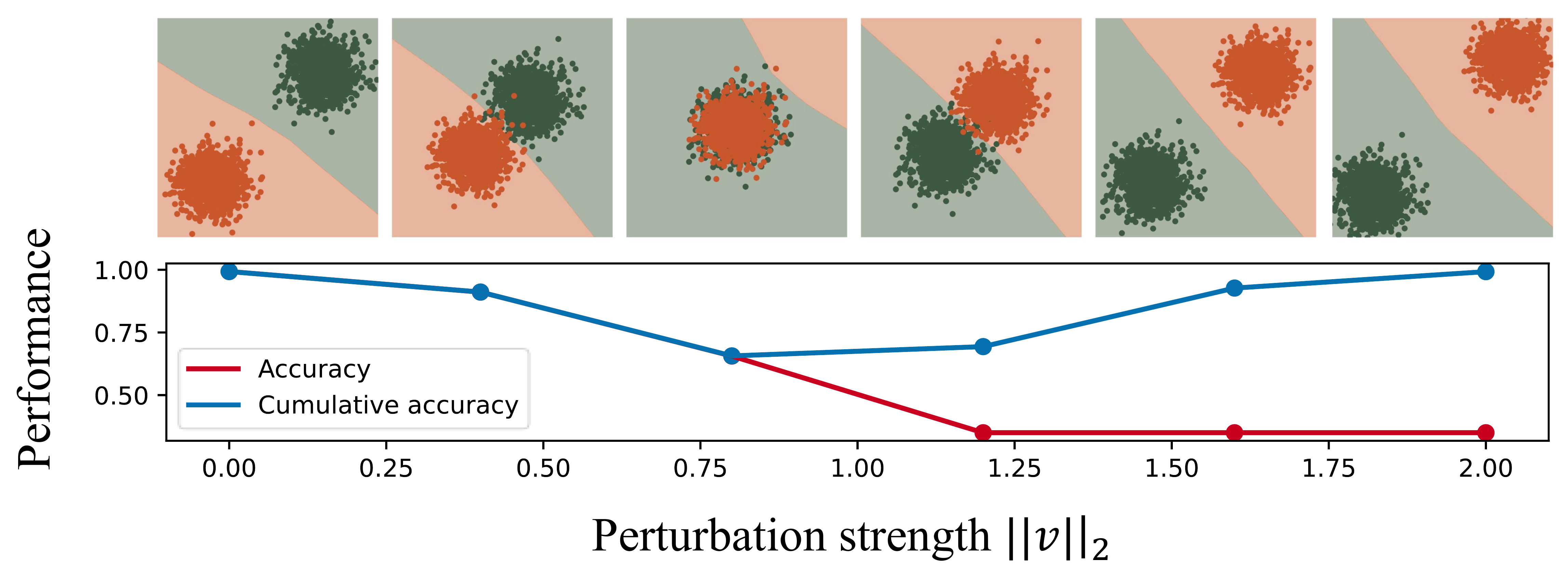}
    \caption{\jjo{\textbf{Monitoring Performance Degradation}.
(top) GOAR perturbs each sample in the opposite feature direction.
Continued perturbation on samples eventually makes mixed samples distinguishable again.
(bottom) Accuracy does not accurately capture the number of samples that lost information, while Cumulative accuracy correctly captures the number of lost samples.}}
    \label{fig:performance_degradation}
\end{figure}



\jo{To summarize, our projection process is as follows: Starting with the perturbed image $x - v$, we introduce a small noise $w$, yielding $\tx = x - v + w$. Next, we employ the diffusion model to denoise $\tx$.} 
This approach effectively eliminates irrelevant information associated with off-manifold directions while keeping the meaningful on-manifold components unchanged. 
We present visual results in \fref{fig:method}.
For details of the hyper-parameter setting and algorithm, please refer to \aref{appendix:implementation_details}, and \aref{appendix:code}, respectively. 


\subsection{\jo{Monitoring Performance Degradation}}

\label{subsection:measure_performance}

ROAR and ROAD measure performance degradation by assessing how much a model's accuracy drops as perturbations are added.
GOAR \jo{adopts a distinct measurement criterion, which involves tallying} the cumulative number of misclassified examples.
\jo{This approach is chosen because, in GOAR, a sample can regain separability after losing its information, effectively becoming indistinguishable from other class labels (\fref{fig:performance_degradation}).}
Therefore, even after the misclassified image becomes distinguishable again, {such cases should be considered as successful instances of performance degradation.}
\subsection{GOAR: Summary}
\label{subsection:summary}
{
In summary, GOAR can be summarized as follows: 
1) Obtain feature vectors $v_i$ corresponding to each sample. 
2) Shift the original image $x_i$ in the direction of $-v_i$ by a certain magnitude. 
3) Using the diffusion model, project the shifted image back onto the image manifold. 
4) Retrain on the modified dataset and count the cumulative number of incorrectly predicted samples.
}

\section{Experiments}

In this section, we aim to validate our analysis and highlight the effectiveness of GOAR through experiments.  
Firstly, we carry out experiments in scenarios where we have ground truth features. 
Comparing the evaluation results based on perturbation based metrics and ground truth features, we verify that GOAR stands out as the only method capable of providing accurate assessments (\sref{subsection:exp_toy_dataset}).
Furthermore, we show that GOAR is applicable across diverse domains including both vision datasets and tabular datasets (\sref{subsection:exp_real_dataset}).
Finally, we demonstrate the necessity of manifold projection through an ablation study (\sref{subsection:exp_ablation_study}).


\paragraph{Experimental setup} 
We use a 3-layer CNN for MNIST \citep{deng2012mnist} and FashionMNIST \citep{xiao2017fashionmnist}, and a ResNet18 \citep{he2016deep} for CIFAR10 \citep{Krizhevsky2009LearningML}. For tabular and synthetic datasets, we utilize a 3-layer MLP.
We compare seven attribution methods: Input gradient (Grad) \citep{simonyan2014deep}, Grad $\times$ Input (I$\times$G) \citep{shrikumar2017just}, SmoothGrad (SG) \citep{smilkov2017smoothgrad}, Integrated gradient (IG) \citep{sundararajan2017axiomatic}, \yh{DeepLift \citep{shrikumar2017learning}, and GradSHAP for image datasets, and KernelSHAP for tabular dataset \citep{lundberg2017unified}.}
\jo{As a control, we opt for a randomly generated vector (Random) as a feature, serving as a lower bound for performance comparison.}
\yh{We use captum library \citep{kokhlikyan2020captum} to implement attribution methods.}
For accurate evaluations, we conduct a retraining for ROAD. 
In image datasets, we compare GOAR with ROAR and ROAD. Since ROAD cannot be used with tabular datasets, we use Eval-X as a competitor instead of ROAD in tabular datasets.
For implementation details, please refer to \aref{appendix:implementation_details}.

\begin{table*}[t]
\caption{{\textbf{Pearson's correlation{s} between performance drop and ground truth evaluation metrics.}  
A stronger correlation indicates that the benchmark aligns more closely with the ground truth evaluation. In almost all cases, it can be observed that GOAR has the highest correlation.
}}
\begin{center}
\begin{small}
\begin{tabular}{c|c|cccccc}
\toprule
{Dataset} & {Benchmark} & FA ($\uparrow$) & RA ($\uparrow$) & SA ($\uparrow$) & SRA ($\uparrow$) & PRA ($\uparrow$) & RC ($\uparrow$) \\
\midrule
\multirow{3}{*}{Adult}      & GOAR        &$\mathbf{+0.97 \pm 0.02}$&$\mathbf{+0.88 \pm 0.11}$&       ${+0.15 \pm 0.20}$&$\mathbf{+0.90 \pm 0.09}$ &$\mathbf{+0.95 \pm 0.03}$&$\mathbf{+0.93 \pm 0.03}$\\
{}                          & ROAR        &       ${-0.75 \pm 0.09}$&       ${-0.37 \pm 0.14}$&       ${-0.77 \pm 0.12}$&       ${-0.42 \pm 0.13}$ &${-0.75 \pm 0.09}$       &       ${-0.84 \pm 0.06}$\\
{}                          & Eval-X      &       ${-0.48 \pm 0.11}$&       ${-0.82 \pm 0.07}$&$\mathbf{+0.75 \pm 0.08}$&       ${-0.79 \pm 0.08}$ &${-0.49 \pm 0.11}$       &       ${-0.34 \pm 0.12}$\\
\midrule
\multirow{3}{*}{Heloc}      & GOAR        &$\mathbf{+0.95 \pm 0.02}$&$\mathbf{+1.00 \pm 0.01}$&$\mathbf{+0.92 \pm 0.03}$&$\mathbf{+1.00 \pm 0.01}$ &$\mathbf{+0.90 \pm 0.03}$&$\mathbf{+0.82 \pm 0.04}$\\
{}                          & ROAR        &       ${-0.63 \pm 0.13}$&       ${-0.41 \pm 0.20}$&       ${-0.68 \pm 0.12}$&       ${-0.42 \pm 0.19}$ &${-0.71 \pm 0.10}$       &       ${-0.80 \pm 0.06}$\\
{}                          & Eval-X      &       ${-0.52 \pm 0.14}$&       ${-0.74 \pm 0.12}$&       ${-0.46 \pm 0.14}$&       ${-0.73 \pm 0.12}$ &${-0.42 \pm 0.14}$       &       ${-0.26 \pm 0.15}$\\
\midrule
\multirow{3}{*}{Synthetic}  & GOAR        &$\mathbf{+0.66 \pm 0.03}$&$\mathbf{+0.98 \pm 0.01}$&$\mathbf{+0.59 \pm 0.03}$&$\mathbf{+0.98 \pm 0.01}$ &$\mathbf{+0.70 \pm 0.03}$&$\mathbf{+0.53 \pm 0.03}$\\
{}                          & ROAR        &       ${-0.87 \pm 0.05}$&       ${-0.44 \pm 0.16}$&       ${-0.91 \pm 0.03}$&       ${-0.43 \pm 0.16}$ &${-0.84 \pm 0.07}$       &       ${-0.92 \pm 0.03}$\\
{}                          & Eval-X      &       ${-0.24 \pm 0.04}$&       ${-0.78 \pm 0.02}$&       ${-0.15 \pm 0.04}$&       ${-0.78 \pm 0.02}$ &${-0.31 \pm 0.04}$       &       ${-0.09 \pm 0.04}$\\
\bottomrule
\end{tabular}
\end{small}
\end{center}
\label{tab:openxai}
\end{table*}

\subsection{\jo{Feature Assessment with Ground Truth Features}}
\label{subsec:synthetic}

In this subsection, we use benchmarks in situations where we have access to the ground truth features. 
\yh{
We conducted experiments in two settings: one where we know the ground truth features of the dataset \cite{xai-bench-2021}, and another where we know the features of the model \cite{agarwal2022openxai}.
We measure how well the benchmark results align with evaluations based on ground truth features, demonstrating that GOAR accurately evaluates the quality of features in both scenarios, while pixel-perturbation strategies fail.
}




\begin{figure}[!t]
    \centering
    \includegraphics[width=1.0\linewidth]{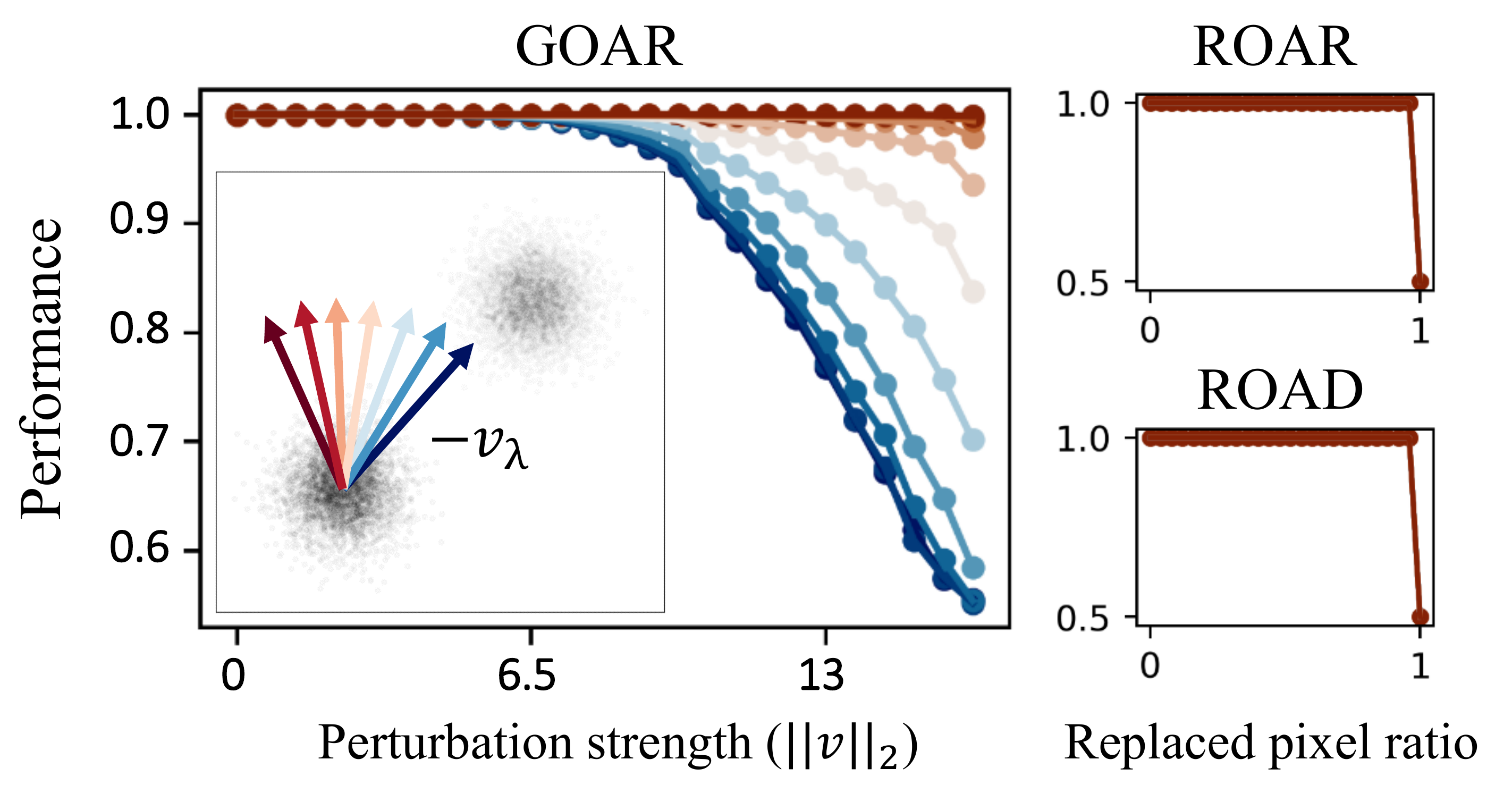}
    \vspace{-0.5em}
    \caption{
    \textbf{Feature assessment with a toy dataset.} 
    Comparison of features using ROAR, ROAD, and GOAR.
    The arrows in inset depict the directions of features with varying magnitudes {$\lambda$} of perturbation.
    {GOAR is the only method where the performance drop becomes significant when the features are similar to the original feature (blue {arrow with $\lambda=1$}).
    \jjo{Note that symbols of different colors overlap in ROAR and ROAD plots.}}
    }
    \label{fig:toy_dataset}
    \vspace{-0.5em}
\end{figure}


\paragraph{Synthetic dataset}
\label{subsection:exp_toy_dataset}

To test whether benchmarks can effectively compare \jo{the quality of features}, we first create a 64-dimensional Gaussian mixture dataset, i.e., $x \sim \mathcal N(x;\mu_k, 0.3\mathbf{I})$. Here, \jo{data $x$, sampled from class 1 and 2, follow $\mu_1 = (1, ..., 1)$ and $\mu_2 = (-1, ..., -1)$, respectively.} When trained on a simple neural network, the input gradients align parallel with the difference of the \jo{$\mu_k$} vectors for each class $k$. 
\jo{In other words, the input gradients effectively capture the ground truth features.}

{We then add noise of varying ratios to the 
input gradient $v$ as $v_{\lambda} = \lambda v + (1-\lambda) w$, where $w$ is sampled from a uniform distribution on the unit sphere, and $\lambda$ is a hyperparameter controlling the noise level.
\fref{fig:toy_dataset}-(a) illustrates $v_\lambda$ with various noise levels, showing that as the noise ratio increases, its direction deviates from the ground truth feature.}

{\fref{fig:toy_dataset} presents the results comparing $v_\lambda$ at different noise levels through ROAR, ROAD, and GOAR.
GOAR shows that when $v_\lambda$ closely resembles the ground truth feature, performance degradation becomes more pronounced and occurs more quickly.
}
However, benchmarks like ROAR and ROAD exhibit no performance difference regardless of noise magnitude.
In essence, while ROAR and ROAD struggle to differentiate between feature qualities, GOAR demonstrates its ability to accurately assess feature quality.
{This is the direct consequence of the pitfall 2, where performance degradation occurs only when all relevant features are removed (\sref{subsection:theory_pitfalls}).}

\begin{figure}[!t]
    \centering
    \includegraphics[width=1.0\linewidth]{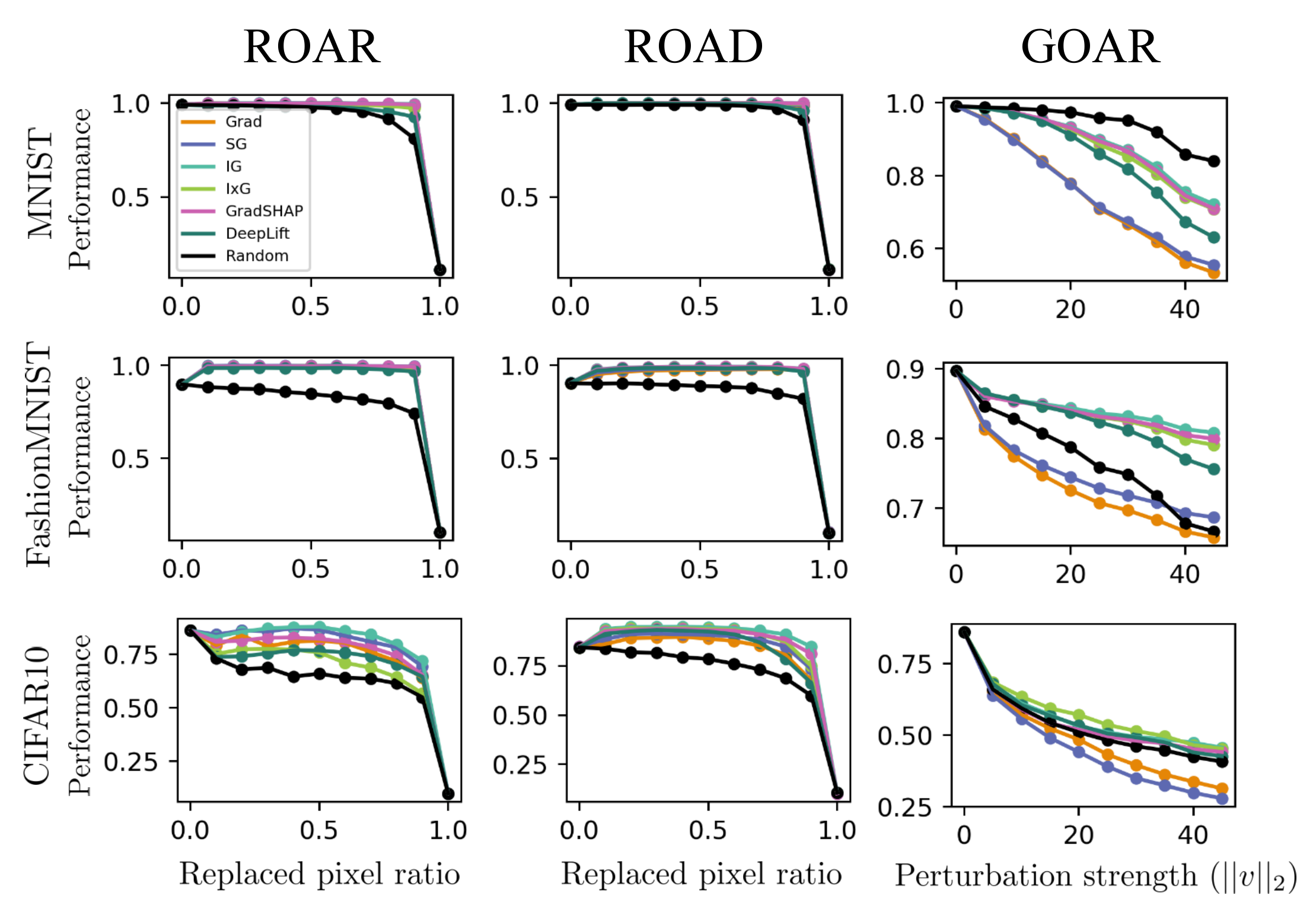}
    \caption{
    \textbf{Feature assessment with vision datasets.} 
    Comparison of attribution methods using ROAR, ROAD, and GOAR. {Real} image datasets of MNIST, FashionMNIST, and CIFAR10 are used for evaluating seven attribution methods.
    }
    \label{fig:vision_dataset}
\end{figure}

\begin{figure}[!t]
    \centering
    \includegraphics[width=1.0\linewidth]{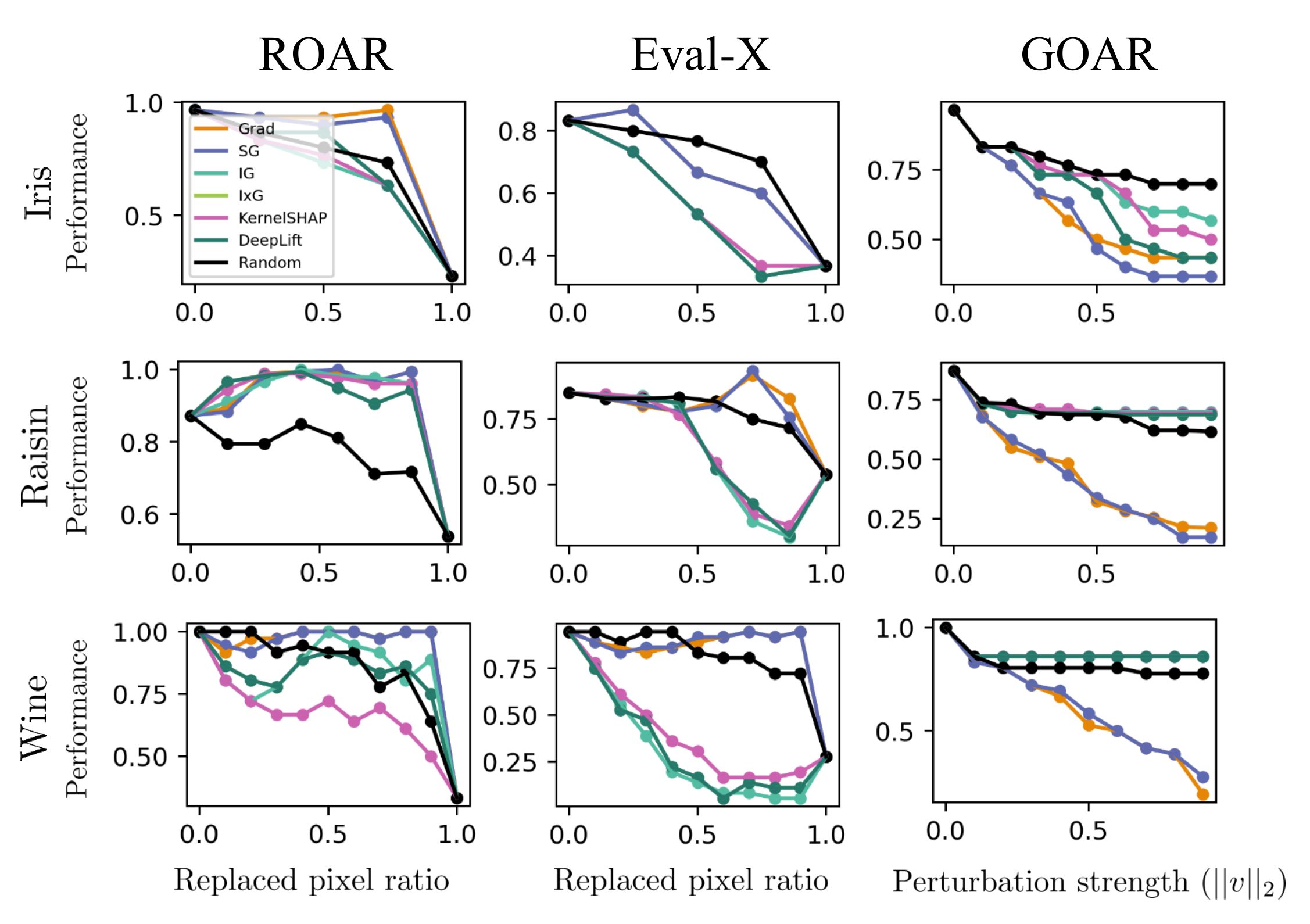}
    \caption{
    \textbf{Feature assessment with tabular datasets.} 
    Comparison of attribution methods using ROAR, Eval-X, and GOAR. {Real} tabular datasets of Iris, Raisin, and Wine are used for evaluating seven attribution methods.
    }
    \label{fig:tabular_dataset}
\end{figure}

\paragraph{OpenXAI benchmark}

The OpenXAI benchmark \citep{agarwal2022openxai} is a metric used to evaluate how similar the features obtained by an attribution method are to the ground truth features, given by an interpretable model, e.g., logistic regression. OpenXAI evaluates attribution methods based on six criteria. 
For example, Feature Agreement (FA) measures the overlap between the top-k pixels discovered through an attribution method and the corresponding ground truth feature.
For more details on other criteria, please refer to the \aref{appendix:openxai}.

\tref{tab:openxai} measures the correlation between the evaluation results of the OpenXAI benchmark and the performance degradation of each benchmark. In almost all cases, it is observed that GOAR exhibits the highest correlation. However, in the case of pixel-perturbation strategies (ROAR, Eval-X), it shows low correlation, sometimes even negative correlation, which raises suspicions about the reliability of the benchmarks.



\subsection{\jo{Feature Assessment with Real Datasets}}


\label{subsection:exp_real_dataset}

\fref{fig:vision_dataset} presents the results of evaluating various attribution methods using ROAR, ROAD, and GOAR. 
ROAR and ROAD exhibit minimal differences in performance within attribution methods, making it nearly impossible to compare different methods, especially on the MNIST and FashionMNIST datasets. 
Furthermore, the fact that all methods perform even worse than the Random, which is a baseline, raises doubts about the results obtained through these approaches. 
On the other hand, GOAR effectively allows the comparison of different methods.

{\fref{fig:tabular_dataset} shows the outcomes of assessing different attribution methods using ROAR, Eval-X, and GOAR.
In the case of GOAR, it demonstrates its capability to efficiently compare various attribution methods, even when dealing with tabular datasets. 
}
Our experiment indicates that Grad and SG consistently exhibit the highest performance, regardless of the specific dataset domain.








\subsection{{Ablation {S}tudy}}

\label{subsection:exp_ablation_study}
\begin{figure}[!t]
    \centering
    \includegraphics[width=1\linewidth]{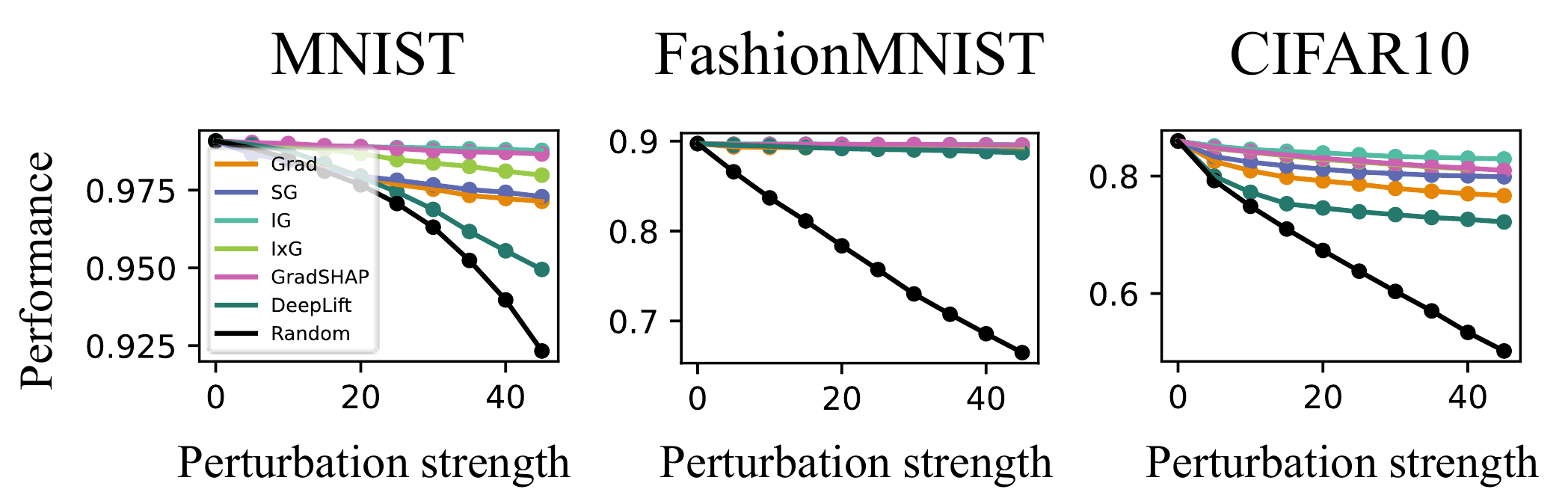}
    \caption{{
    \textbf{Feature assessment {of} GOAR without manifold projection.} 
    {In the absence of} manifold projection, GOAR fails to determine whether attribution methods outperform {the baseline of Random}, akin to the challenge faced by ROAR and ROAD. This issue arises from the class information leakage, which hinders the quantification of removed information.
    }}
    \label{fig:ablation_no_adv}
\end{figure}

In \fref{fig:ablation_no_adv}, we present the results of the GOAR experiment conducted without the manifold projection. 
In contrast to when the projection is employed, the performance of each attribution method is not distinctly discernible. 
Moreover, all the attribution methods are rated as underperforming compared to the Random baseline, which presents a significant problem.

\begin{figure}[!t]
    \centering
    \includegraphics[width=1\linewidth]{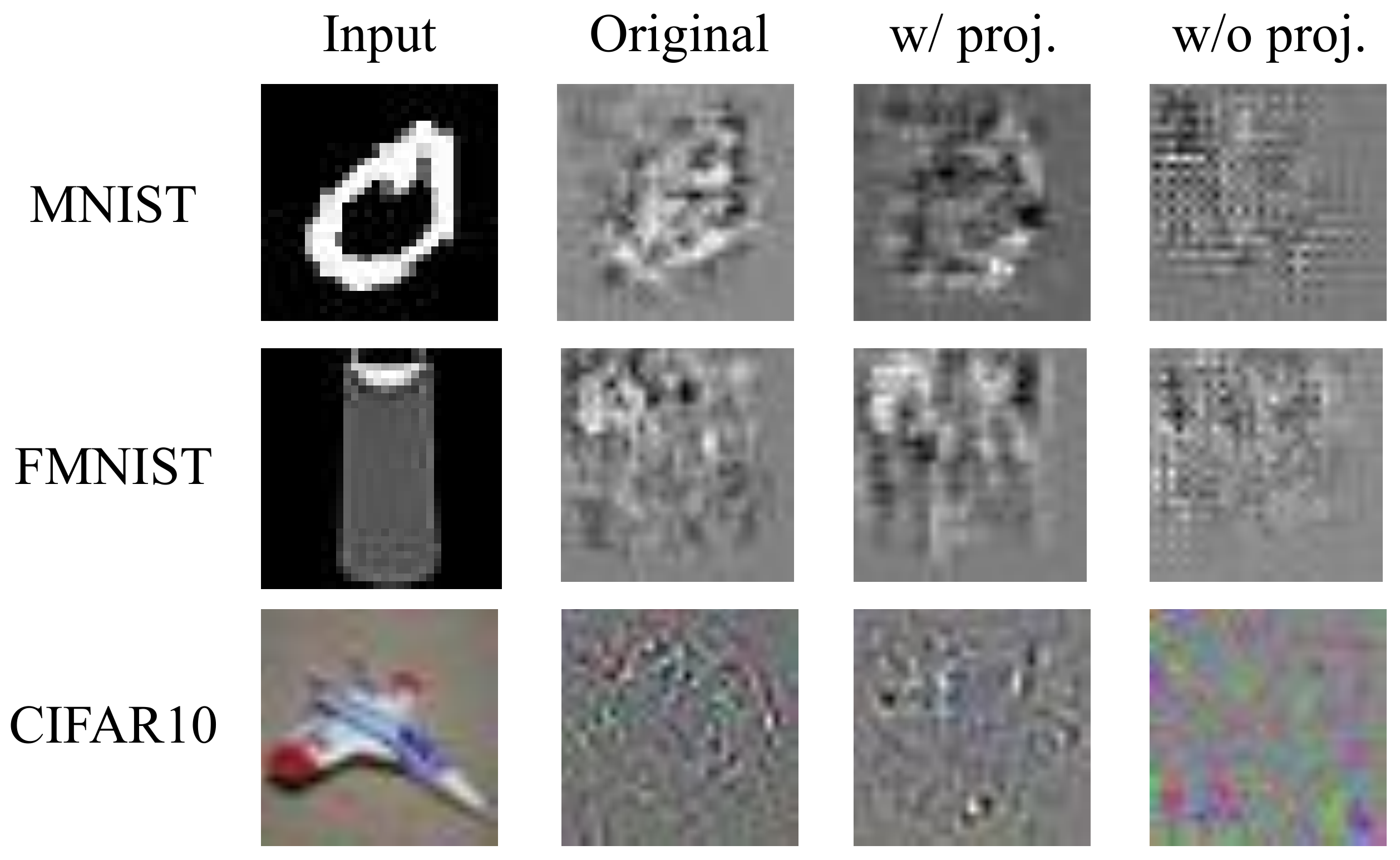}
    \caption{{
    \textbf{Saliency map of the retrained model {with} GOAR.} 
    ``Original'' represents the input gradient of the model trained on the original dataset. ``w/ proj.'' and ``w/o proj.'' represent the input gradient of the model trained on the modified dataset during GOAR, with and without manifold projection, respectively.
    While Grad and w/ proj. display similar features, w/o proj. exhibits spurious patterns.
    }}
    \label{fig:ablation_no_adv_obs}
\end{figure}

This issue stems from the class information leakage, where the retrained model learns unintended, spurious off-manifold components in the modified image $\tx$.
We verify this by examining the saliency maps of retrained models. 
In \fref{fig:ablation_no_adv_obs},
it is clear that a model trained on a modified dataset during GOAR without projection (\textit{w/o proj.}) exhibits entirely different features, such as grid-like patterns.
These results demonstrate the necessity of manifold projection to prevent class information leakage.

\section{Discussions}
\label{sec:discussions}
\jo{GOAR overcomes the constraints of pixel-perturbation strategies, but it still has limitations stemming from its reliance on the diffusion model.}
\jo{First, employing the diffusion model is time-consuming.} 
For specific \jo{computational} costs, \aref{appendix:implementation_details} \jo{can be referred.} 
Additionally, when features have low-frequency components in their off-manifold direction, the diffusion model may struggle to effectively project the modified image onto the data manifold \citep{zhao2022towards}.
Therefore, future research should focus on debiasing the retrained model without relying on \jo{the} diffusion model.

{
Our analysis highlights the dependency of ROAR on a pixel basis as a concern. However, if the primary objective of the evaluation is not to identify meaningful feature vectors within the dataset but, instead, to rank the importance of individual pixels in an image, then erasing based on a pixel basis can be considered acceptable. Nonetheless, it is still crucial to understand that the pixel-perturbation strategy does not discriminate between differences in features at relevant coordinates (pitfall 2, \sref{subsection:exp_toy_dataset}).
}

\section{Conclusion}

In conclusion, we analyzed pixel-perturbation metrics like ROAR, Eval-X, and ROAD from a geometric perspective, revealing their limitations in comparing feature attribution methods.
Their approaches are coordinate-dependent and fail to discriminate differences in features at relevant coordinates. 
To address these issues, we introduced a novel feature-perturbation metric \jo{named} GOAR. 
With experiments on both toy and real datasets, we demonstrated that GOAR outperforms pixel-based approaches in comparing features.
We hope our geometric analysis and metric help cultivate a new insight into feature attribution methods and their evaluation metrics. 

\section{Impact Statements}

This paper presents work whose goal is to advance the field of Machine Learning. There are many potential societal consequences of our work, none of which we feel must be specifically highlighted here.


\bibliography{reference_papers}
\bibliographystyle{icml2024}

\newpage
\appendix
\onecolumn
\section{Implementation Details}
\label{appendix:implementation_details}

\paragraph{Classification model} 
For experiments with the MNIST and FashionMNIST, we employ a 3-layer CNN architecture. Each layer has channel numbers 32, 64, and 64, with kernel sizes of 5, 3, and 3, respectively. After the CNN layers, the output went through two fully connected (FC) layers. These FC layers had widths of 64 and 32, respectively. The hyperparameters used for training were as follows: Adam optimizer, learning rate $= 3 \times 10^{-4}$, batch size $= 256$, and early stopping criteria $= 5$.

For experiments with the CIFAR10, we used a ResNet18 architecture. The hyperparameters used for training were as follows: SGD optimizer, epochs $= 200$, learning rate $= 10^{-1}$ with a cosine learning rate schedule, batch size $= 128$, weight decay $5\times 10^{-4}$, and early stopping criteria $= 5$.

For experiments with tabular and synthetic datasets, we employ a 3-layer MLP architecture. Each layer has width of 128.

\paragraph{Diffusion model} 
For all experiments, we used the DDPM \citep{ho2020denoising} model.
We use the DDIM \citep{song2021denoising} sampler for inference. 
We set the total number of inference steps $=25$. 

\paragraph{Manifold projection} 
To effectively remove the off-manifold component in $v$, we add additional noise $w$ to perturbed image $x-v$.
We add noise of a magnitude that increases the diffusion timestep by $0.16T$, where $T$ is a maximum diffusion timestep, i.e. $1000$.
It corresponds to the additional DDIM inference step of 4.

\paragraph{Computational cost} 
High computational cost is one of the significant limitations of GOAR. 
For the most time-consuming experiment involving the CIFAR10 dataset, the process of manifold projection for the entire dataset typically takes approximately 3 to 4 minutes when using an Nvidia RTX 3090. As a result, there is a need for future research to find ways to either eliminate the use of the diffusion model or decrease the computational demands to make it more efficient.

\section{OpenXAI Benchmark}
\label{appendix:openxai}

In this section, we describe six evaluation metrics utilizing ground truth features in the OpenXAI benchmark. 

\paragraph{Feature Agreement (FA)} Feature Agreement measures the overlap ratio between the top-k pixels found through XAI methods and the top-k pixels found based on ground truth features.

\paragraph{Rank Agreement (RA)} Rank Agreement measures the correlation between the rank order of the top-k pixels found through the XAI method and the rank order of the top-k pixels found based on ground truth features.

\paragraph{Sign Agreement (SA)} Sign Agreement compares the signs of the top-k pixels found through the XAI method with the signs of the top-k pixels found based on ground truth features.

\paragraph{Signed Rank Agreement (SRA)} Signed Rank Agreement compares the signs and rank order of the top-k pixels found through the XAI method with the signs and rank order of the top-k pixels found based on ground truth features.

\paragraph{Rank Correlation (RC)} Rank Correlation calculates Spearman's rank correlation coefficient between the feature ranking order found through the XAI method and the ground truth feature ranking order.

\paragraph{Pairwise Rank Agreement (PRA)} Pairwise Rank Agreement compares the pairwise ranking of features found through the XAI method with the pairwise ranking of ground truth features.

\section{Pitfalls of pixel-perturbations}

\label{appendix:pitfalls}

In this section, we discuss how Eval-X and ROAD also share geometric limitations similar to those of ROAR. 
In our work, we only verify the pitfalls of pixel-perturbation strategies in the simplest scenario. Expanding this result to a generalized setting is an important future work.

\paragraph{Eval-X}

Let's assume that we train a proxy model through an infinite amount of computation. In that case, the modified dataset by ROAR becomes a part of Eval-X's training dataset. Therefore, if we train for a sufficient amount of time, Eval-X becomes a good proxy for the retrained model in ROAR, as Eval-X assumes. However, it means that Eval-X's proxy model also suffers from the same issues as ROAR stemming from the pixel-perturbation strategy.

\begin{figure*}[!t]
    \centering
\includegraphics[width=1.0\linewidth]{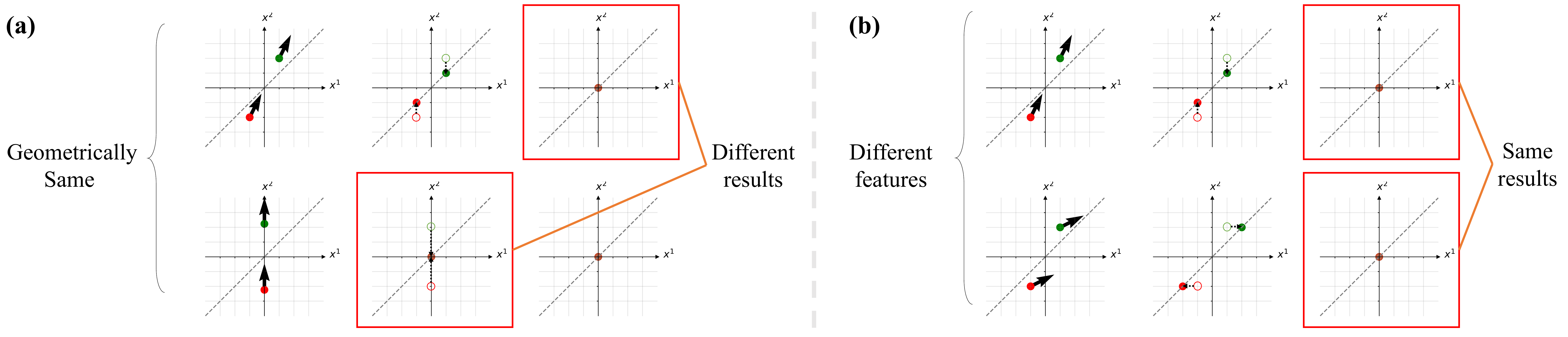}
    \caption{\textbf{Pitfalls of ROAD.}
    Red and Green circles are two data points corresponding to different classes. Black arrows ($\rightarrow$) represent the feature vectors obtained from the attribution method and dashed arrows ($\dashedrightarrow$) 
    indicate the pixel-perturbation based on ROAD. 
    Red borders indicate the points where the performance drop occurs.
    (a) ROAD produces different results in the same geometrical situation depending on coordinate choices.
    (b) ROAD can produce the same results with different features, which makes it challenging to discriminate between these various features.
    }
    \vspace{-1em}
    \label{fig:pitfalls_road}
\end{figure*}

\paragraph{ROAD}

By applying the same line of reasoning discussed in \sref{subsection:theory_pitfalls} to ROAD, we show that ROAD also faces the same problems as ROAR. \fref{fig:pitfalls_road} summarizes the overall discussions.

Here, we consider $\mathcal{D}={(x_1, 0), (x_2, 1)}$ with only one data point for each of the binary labels, 0 or 1. Let's denote the difference vector between the two samples as $dx \equiv x_1 - x_2$. Furthermore, $v_1$ and $v_2$ are the features obtained through the attribution method for each sample $x_1$ and $x_2$, respectively. For simplicity, let's assume that $dx = v_1 = v_2$.

First, ROAD is not invariant to coordinate transformations (pitfalls 1, \fref{fig:pitfalls_road}-(a)).
Let's consider a different coordinate system where we represent the same dataset with $dx = (dx_1, 0, \cdots, 0)$. In this new coordinate system, the feature representation becomes $v_1 = v_2 = (1, 0,\cdots, 0)$. As a result, removing just one pixel in this coordinate system leads to performance degradation.

Second, ROAD is not discriminative to differences in features at relevant coordinates. (pitfalls 2)
To illustrate this, \jo{we} consider a dataset $\mathcal D$ defined by $dx = (1, 2, \epsilon)$, 
\jo{and} two attribution methods, denoted as $e$ and $e'$.
Let's assume features derived from $e$ are $v_1 = v_2 = (1, 2, 0)$, while those from $e'$ are $v'_1 = v'_2 = (2, 1, 0)$.
Ideally, one might anticipate ROAD to favor method $e$ due to its feature alignment with the distinctive direction, i.e., $dx$, of the samples.
However, as shown in \fref{fig:pitfalls_road}-(b), both $e$ and $e'$ witness an accuracy drop after two pixels are removed, rendering them indistinguishable.
This example highlights that variations in relevant coordinates do not influence ROAD's outcomes, making it unable to distinguish between methods $e$ and $e'$.

\section{Algorithm}
\label{appendix:code}

In this section, for reproducibility, we provide the PyTorch \cite{paszke2017automatic} code for manifold projection.
The source code for our experiments will be publicly available upon publication.

First, we determine the diffusion timestep $t_{x-v}$ that matches the magnitude of the added $v$.
Next, we add noise $w$ of the predetermined size to $x$ and adjust the timestep accordingly. In this case, the timestep becomes $t_{x-v+w} = t_{x-v} + 0.16T$.
Finally, we sample $x-v+w$ with DDIM from $t_{x-v+w}$ to 0.

\begin{lstlisting}[language=Python, caption=Python example, caption={\textbf{Manifold projection}}, captionpos=b]
import torch

def perturbation_strength_to_t(perturbation_strength, scheduler, in_ch=3, diffusion_timestep=1000):
    '''
    Args
        perturbation_strength : l2 norm of v
        scheduler : huggingface diffusers scheduler
        in_ch : number of input channels
    '''
    noise_to_signal_ratio = (1-scheduler.alphas_cumprod).sqrt() / scheduler.alphas_cumprod.sqrt()
    dnsr = noise_to_signal_ratio - np.abs(remove_size) * self.init_trans.std.mean() / 0.5 / np.sqrt(in_ch*32*32)
    t    = dnsr.abs().argmin() / diffusion_timestep
    return t

def t_to_t_idx(t, scheduler, diffusion_steps=1000):
    return (scheduler.timesteps - diffusion_steps*t).abs().argmin()

def adv_purification(x, diffusion_model, scheduler, perturbation_strength, additional_noise_t_idx=4):
    '''
    Args
        x : perturbed image, i.e., x - v
        diffusion_model : pretrained diffusion model
        perturbation_strength : norm of v
        scheduler : DDIM scheduler, you could find it in Huggingface diffusers library
        additional_noise_t_idx : the size of additional noise for better purification
    '''
    # fix random seed
    generator = torch.Generator().manual_seed(SEED)
    
    # define timesteps
    scheduler.set_timesteps(num_inferences=25)

    # estimate proper diffusion timestep for x-v+w
    t_x = perturbation_strength_to_t(remove_size, scheduler)
    t_idx_x = t_to_t_idx(t_x, scheduler)
    t_idx_x_plus_noise = t_idx_x+additional_noise_t_idx
    t_x_plus_noise = scheduler.timesteps[t_idx_x_plus_noise]

    # add additional noise for better adversarial purification
    noise_signal_ratio = (1-scheduler.alphas_cumprod).sqrt() / scheduler.alphas_cumprod.sqrt()
    xt = x + (noise_signal_ratio[t_x_plus_noise] - noise_signal_ratio[t_x]) * torch.randn(x0.shape, generator=generator)
    xt = torch.sqrt(self.alphas_cumprod[t_x_plus_noise]) * xt

    # DDIM sampling (xt -> x0)
    for t in enumerate(scheduler.timesteps):
        if t > t_x_plus_noise:
            continue
        et = diffusion_model(xt, t)
        xt = scheduler.step(et, t, xt, eta=0, generator=generator).prev_sample
    x0 = xt
    
    return x0
\end{lstlisting}


\end{document}